%% file: main.tex
\documentclass[runningheads]{llncs}

\usepackage{eccv}

\usepackage{eccvabbrv}

\usepackage{graphicx}
\usepackage{booktabs}

\usepackage[accsupp]{axessibility}  

\usepackage{orcidlink}

\usepackage{bm}
\usepackage{multirow}

\makeatletter
\newcommand{\printfnsymbol}[1]{%
  \textsuperscript{\@fnsymbol{#1}}%
}
\makeatother

\begin{document}

\title{A Lost Opportunity for Vision-Language Models: A Comparative Study of Online Test-Time Adaptation for Vision-Language Models}

\titlerunning{Online Test-Time Adaptation for Vision-Language Models}

\author{%
Mario D\"obler\thanks{Equal contribution.} \and
Robert A. Marsden\printfnsymbol{1} \and
Tobias Raichle \and
Bin Yang
}

\authorrunning{M.~D\"obler et al.}

\institute{University of Stuttgart, Germany\\
\email{\{mario.doebler, robert.marsden, tobias.raichle, bin.yang\}@iss.uni-stuttgart.de}}

\maketitle

\begin{abstract}
In deep learning, maintaining model robustness against distribution shifts is critical. This work explores a broad range of possibilities to adapt vision-language foundation models at test-time, with a particular emphasis on CLIP \cite{radford2021learning} and its variants. The study systematically examines prompt-based techniques and existing test-time adaptation methods, aiming to improve the robustness under distribution shift in diverse real-world scenarios. Specifically, the investigation covers various prompt engineering strategies, including handcrafted prompts, prompt ensembles, and prompt learning techniques. Additionally, we introduce a vision-text-space ensemble that substantially enhances average performance compared to text-space-only ensembles. Since online test-time adaptation has shown to be effective to mitigate performance drops under distribution shift, the study extends its scope to evaluate the effectiveness of existing test-time adaptation methods that were originally designed for vision-only classification models. Through extensive experimental evaluations conducted across multiple datasets and diverse model architectures, the research demonstrates the effectiveness of these adaptation strategies. Code is available at: \url{https://github.com/mariodoebler/test-time-adaptation}
\end{abstract}

\section{Introduction}
In the rapidly evolving field of deep learning, the robustness of models against distribution shifts remains a critical challenge. If the data distribution at test-time deviates from the training distribution, the performance can decrease significantly. This challenge is prevalent in most practical deep learning applications due to the difficulty of accurately replicating testing conditions during training. An intuitive answer to shifts in distribution is an extensively trained model across diverse datasets that can be adapted to a wide range of downstream tasks. Such models are nowadays termed as foundation models. They are known to exhibit superior generalization abilities, setting them apart from conventional models. Current vision-language models, like CLIP \cite{radford2021learning}, have shown strong zero-shot performance across a variety of computer vision benchmarks.

In this work, we study the task of online test-time adaptation (TTA) for vision-language (VL) models, with a specific focus on CLIP and its variants. We explore various strategies and methodologies aimed at enabling these models to adapt dynamically to distribution shifts encountered during inference. Our investigation encompasses both prompt-based approaches, which involve modifying the input prompts provided to the model, and existing TTA methods borrowed from the domain of image classification. By systematically evaluating these approaches across a range of datasets and scenarios, we aim to provide insights into the efficacy and practical applicability of different TTA strategies for vision-language models. Through our exploration, we seek to contribute to the development of more robust and adaptable vision-language models capable of performing reliably in diverse real-world settings.

We summarize our main contributions as follows:
\begin{itemize}
    \item We discuss a broad range of possibilities to adapt vision-language foundation models at test-time - from various prompting strategies to applying existing test-time adaptation methods.
    \item We introduce a vision-text-space ensemble that is optimization-free and outperforms test-time prompt tuning.
    \item Our broad comparative study shows the potential of existing test-time adaptation methods for enhancing the robustness of vision-language models. Choosing a good method leads to significant improvements across a braod range of datasets and models.
\end{itemize}

\section{Related Work}
\subsection{Foundation Models}
''Foundation model'' is a general notion of systems with broad zero-shot capabilities that can be adapted for specific purposes, e.g., via fine-tuning. Most notably, this encompasses large language models (LLMs) and multimodal models, such as large vision-language models (VLMs). LLMs are systems capable of understanding and generating language; popular examples include \cite{brown2020language, devlin2018bert, touvron2023llama}. VLMs combine visual and textual information, enabling them to comprehend and generate content that encompasses both modalities. Several VLM architectures have been proposed: dual-encoder architectures \cite{radford2021learning, jia2021scaling}, encoder-decoder architectures \cite{wang2021simvlm, chen2022pali}, unified transformer architectures \cite{li2022blip, bao2021beit}, and many more. In this work we investigate test-time adaptation for VLMs and mainly focus on CLIP \cite{radford2021learning}, as it is still the most representative VLM. Additionally, we report results for EVA-CLIP \cite{sun2023eva} that proposed improved training techniques for CLIP at scale.

\subsection{Online Test-Time Adaptation}
Online test-time adaptation adapts the model to an unknown domain shift directly during inference, leveraging all available test samples. One successful line of work recalculates the batch normalization (BN) statistics during test-time \cite{schneider2020improving} to mitigate covariate shift caused by corruption. Although updating only the BN statistics is computationally efficient, it has its limitations, especially regarding natural domain shifts. As a result, recent TTA methods additionally incorporate model weight updates through self-training. TENT \cite{wang2021tent}, for example, showcased that minimizing the entropy with respect to the batch normalization parameters can successfully improve the performance for single-target adaptation. Building upon this idea, EATA \cite{niu2022efficient} introduces a loss weighting and filtering scheme that accounts for the reliability and diversity of a sample. Furthermore, they use elastic weight consolidation \cite{kirkpatrick2017overcoming} to mitigate catastrophic forgetting \cite{mccloskey1989catastrophic} on the initial training domain. However, accessing data from the initial training domain may not always be feasible in practical scenarios. To prevent a model from collapsing to trivial solutions induced by confidence maximization, \cite{liang2020we, mummadi2021test} apply diversity regularizers. Other works, such as \cite{chen2022contrastive, dobler2023robust} also employ contrastive learning to mitigate a domain shift during test-time.

While certain TTA methods focus solely on adapting to a single domain, real-world scenarios often involve encountering multiple domain shifts. Thus, \cite{wang2022continual} introduced continual test-time adaptation, where a model is adapted to a sequence of diverse domains. While self-training-based approaches such as \cite{wang2021tent} can be also utilized in the continual setting, they can be susceptible to error accumulation \cite{wang2022continual, marsden2024universal}. To address this, \cite{wang2022continual} proposes weight and augmentation-averaged predictions alongside a stochastic restore mechanism to mitigate catastrophic forgetting. RMT \cite{dobler2023robust} proposes a robust mean teacher to handle multiple domain shifts, while GTTA \cite{GTTA} uses mixup and style-transfer to artificially create intermediate domains. 

Recent research has tackled even more challenging scenarios, such as dealing with temporally correlated data. LAME \cite{boudiaf2022parameter} focuses on adapting the model's output using Laplacian adjusted maximum-likelihood estimation. On the other hand, NOTE \cite{gong2022note}, RoTTA \cite{yuan2023robust}, and DAB \cite{dobler2024diversity} introduce a buffer to simulate an i.i.d. test stream. To handle large and noisy gradients that can promote trivial solutions, SAR \cite{niu2023towards} proposes a sharpness-aware and reliable entropy minimization method. Building upon SAR, DeYO \cite{lee2024entropy} incorporates a confidence metric that measures the extent to which the probability of pseudo-label decreases after applying an image transformation that distorts the shape of the objects.

In the work of \cite{marsden2024universal}, recent TTA methods are evaluated on a broad range of possible TTA scenarios, termed Universal TTA. Their proposed method ROID \cite{marsden2024universal} puts emphasis on using certainty and diversity weighting to prevent the occurrence of trivial solutions during the adaptation. To further preserve the model's generalization capabilities and overcome catastrophic forgetting, ROID introduces weight ensembling. This approach continuously combines the weights of initial source model with those of the current adaptation model during test-time. CMF \cite{lee2024continual} builds upon the ROID framework and replaces weight ensembling by continual momentum filtering. It utilizes a Kalman filter to derive a model that is both resilient to catastrophic forgetting and highly adaptable.

Due to their multimodality and zero-shot generalization capabilities, VLMs offer new possibilities for TTA. One approach that focuses on adapting the prompt space is TPT \cite{shu2022test}. It is inspired by the supervised context optimization (CoOp) \cite{zhou2022learning} approach, but differs by directly optimizing the prompt context during test-time. This is achieved by minimizing the entropy of an augmented batch generated from a single test sample. In this work, we aim to provide new perspectives on how to deal with VLMs, namely CLIP, in the context of online test-time adaptation.

\section{Prompts and Vision-Text-Space Ensembles}
In this chapter, we first revisit the underlying principles of vision-language foundation models such as CLIP, along with their approach to perform zero-shot classification. We then introduce improved prompting strategies, including our novel approach, and provide a comprehensive benchmark of all methods. Building upon this foundation, Chapter 4 explores the combination of improved prompting techniques with existing TTA methods that update the model parameters.

Vision-language foundation models, in particular CLIP \cite{radford2021learning}, aim to learn a joint embedding space for the vision and language modality. This is achieved by aligning the representations of images and their associated textual descriptions through contrastive learning. To extract the embeddings, CLIP leverages a separate encoder for the vision and text modality, denoted here as $f_{\mathrm{vision}}$ and $f_{\mathrm{text}}$, respectively. After successfully training the encoders on typically hundreds of millions of image-text pairs, the learned joint embedding space allows to associate similar concepts across modalities, resulting in cross-modal understanding. 

To perform zero-shot classification with a hand-crafted prompt, the procedure involves the following steps. Let $\bm{x}_t \in \mathbb{R}^{H \times W \times C}$ be the current test image at time step $t$ with height $H$, width $W$, and $C$ channels, and $\bm{z}_t = f_{\mathrm{vision}}(\bm{x}_t)$ denote its corresponding representation. In addition, let $\{\bm{t}_k\}_{k=1}^K$ be a textual representation for each of the $K$ classes, obtained by embedding short phrases (templates) like "\textit{a photo of a \{classname\}.}" into the text embedding space. Now, to determine the class label for an image, its representation $\bm{z}_t$ is first paired with each of the $K$ text representations $(\bm{t}_k, \bm{z}_t)$. Then, the cosine similarity $s_k = \mathrm{sim}(\bm{t}_k, \bm{z}_t)$ is computed for each pair. The final model prediction simply corresponds to the class with the highest similarity score or highest softmax probability. The latter can be computed with 
\begin{equation}
    p_{tk} = \frac{\exp(\mathrm{sim}(\bm{t}_k, \bm{z}_t) / \tau)}{ \sum_{j=1}^K \exp(\mathrm{sim}(\bm{t}_j, \bm{z}_t) / \tau)}
    \label{eq:softmax_probs}
\end{equation}
where $\tau$ is a temperature. Note that the text embeddings $\{\bm{t}_k\}_{k=1}^K$ are typically precomputed once before inference, ensuring efficiency during test-time.

\subsection{Prompt Engineering}
While using a simple phrase like \textit{a photo of a \{classname\}.} can already work exceptionally well, the performance of VL models heavily depends on the utilized prompt and its encoded representation \cite{zhou2022learning}. Thus, writing better hand-crafted prompts can significantly improve the performance of the model. For instance, in the context of ImageNet-R \cite{hendrycks2021many} and a ViT-B-16 model pretrained by OpenAI, employing the prompt template "\textit{depiction of a \{classname\}.}" reduces the error rate from 26.0\% to 23.6\%. Similarly, for datasets like EuroSAT \cite{eurosat} that contain low-resolution satellite photos, using a prompt such as "\textit{a blurry satellite photo of \{classname\}.}" decreases the error rate from 58.5\% to 46.3\%. These examples underscore the importance of well-designed prompts to maximize performance.

Instead of relying on a single prompt template, Radford et al. \cite{radford2021learning} also proposed to use a list of $J$ different templates. An example can look like the following list ["\textit{a photo of a \{classname\}}.", "\textit{a sketch of a \{classname\}}.", "\textit{a painting of a \{classname\}."}]. By averaging the text representations obtained from all templates for class $k$, i.e., 
\begin{equation}
    \bar{\bm{t}}_k = \frac{1}{J} \sum_{j=1}^J \bm{t}_{kj},
\end{equation}
a text-based ensemble within the embedding space can be formed. In this case, the similarity scores are now computed with $s_k = \mathrm{sim}(\bar{\bm{t}}_k, \bm{z}_t)$. While this has been found to not only consistently improve the results \cite{radford2021learning}, it also avoids increasing the computational complexity and the memory requirements during inference. This efficiency is again due to the ability of precomputing $\bar{\bm{t}}_k$ prior to inference.

While the ensemble approach described earlier uses a predefined list of hand-crafted prompt templates, CuPL \cite{pratt2023does} introduces a novel strategy that harnesses the power of a large language model to generate a class-specific prompt list. Specifically, the LLM is asked to write descriptive sentences that encapsulate the discriminative features of the various classes. In the case of the category goldfish, the prompt list might look like ["\textit{Most goldfish have a shiny gold or orange color.}", "\textit{A goldfish in a bowl.}", \dots]. These descriptive prompts help to improve the performance of the VL model without requiring any expert knowledge.

\subsection{Learning Prompts}
Zhou et al. \cite{zhou2022learning} introduced context optimization to offline fine-tune CLIP-like vision-language models with a few labeled training examples $\{\bm{x}_i, \bm{y}_i\}_{i=1}^N$, where $\bm{y}_i \in \mathbb{R}^K$ is the one-hot encoded category of image $\bm{x}_i$. Unlike before, where the context of the prompt (such as "\textit{a photo of a}") was either fixed or manually tuned, it is now learnable. This involves representing the context with a few learnable token embeddings, which are then optimized by minimizing, for example, a cross-entropy (CE) loss according to
\begin{equation}
    \mathcal{L}_{\mathrm{CE}} = - \frac{1}{N} \sum_{i=1}^N \sum_{k=1}^K y_{ik} \log(p_{ik}).
\end{equation}
Building upon the idea of learning the context prompts, TPT \cite{shu2022test} exploits context optimization during test-time. The procedure involves using test-time augmentation to create a batch $\{\bm{x}_{ti}\}_{i=1}^B$ of $B=64$ samples from a single test image. Then, the most confident $\rho = 10\%$ of the samples in terms of entropy $e_{ti} = \sum_{k=1}^K p_{tik} \log(p_{tik})$ are selected to minimize an entropy loss with respect to the trainable context parameters. This results in the following expression
\begin{equation}
    \mathcal{L}_{\mathrm{TPT}} = - \frac{1}{\rho B} \sum_{i=1}^B \sum_{k=1}^K [e_{ti} \leq \beta] p_{tik} \log(p_{tik}),
\end{equation}
where $[ \cdot ]$ is the Iverson bracket and $\beta$ is a threshold. After the context is updated one (or several) times, regular zero-shot classification can be performed. While \cite{shu2022test} demonstrate the effectiveness of this approach, it incurs substantial computational overhead. Specifically, each test image results in 64 forward passes through the image encoder and at least 2 forward passes through the text encoder - one for learning an improved context and another to acquire the new text representations.

\subsection{Vision-Text-Space Ensemble}
\input{figures/overview}
The effectiveness of methods like TPT depends on the creation of suitable training examples via test-time augmentation, which can subsequently be identified with confidence-based filtering, for example. Feng et al. \cite{feng2023diverse} take this approach one step further by additionally exploiting Stable Diffusion \cite{rombach2022high} to generate new images that resemble the current test image. Since leveraging a diverse set of augmented test samples might also be helpful for the previous hand-crafted or LLM-based prompt ensemble approaches, we introduce a \textbf{V}ision-\textbf{T}ext-Space \textbf{E}nsemble (VTE), which creates an ensemble in both spaces. Following TPT, VTE utilizes the same test-time augmentation strategy and entropy-based confidence filtering to extract reliable samples from the artificially generated batch. The representations of the identified samples are then averaged via the equation
\begin{equation}
    \bar{\bm{z}}_t = \frac{1}{\rho B} \sum_{i=1}^B [e_{ti} \leq \beta] \bm{z}_{ti},    
\end{equation}
where $\bar{\bm{z}}_t$ is subsequently utilized to compute the similarity scores according to $s_k = \mathrm{sim}(\bar{\bm{t}}_k, \bar{\bm{z}}_t)$. Note that this procedure is again optimization-free and, unlike TPT, does not require any forward passes through the text encoder during test-time. An illustration of VTE is  also shown in \cref{fig:settings}.

\subsection{Experiments}
\subsubsection{Datasets, Models, and Metric}
We follow the continual test-time adaptation setting in \cite{marsden2024universal} and evaluate the models' robustness on ImageNet (validation set) and its variants. ImageNet-C \cite{hendrycks2019benchmarking} includes 15 types of corruptions with 5 severity levels applied to the validation images of ImageNet (IN). For the natural domain shifts, we consider ImageNet-R \cite{hendrycks2021many}, ImageNet-Sketch \cite{wang2019learning}, as well as ImageNet-D109, a variation of ImageNet-D \cite{rusak2022imagenet} introduced in \cite{marsden2024universal}. While ImageNet-R contains 30,000 examples depicting different renditions of 200 IN classes, ImageNet-Sketch contains 50 sketches for each of the 1,000 IN classes. Additionally, we report results for ImageNet-V2 \cite{recht2019imagenet} and ImageNet-A \cite{hendrycks2021natural}. ImageNet-V2 is an independent test set containing 10,000 natural images covering all 1,000 IN classes. ImageNet-A comprises 7,500 adversarial examples for a subset of 200 IN classes.

To evaluate categories outside the ImageNet context, we follow \cite{shu2022test} and report results for ten datasets, covering fine-grained classifications including species of plants or animals (Flowers102~\cite{flowers102}, OxfordPets~\cite{oxfordpets}), scenes \\ (SUN397~\cite{sun397}), textures (DTD~\cite{DTD}), food (Food101~\cite{food101}), transportation (StanfordCars~\cite{stanfordcars}, Aircraft~\cite{aircrafts}), human actions (UCF101~\cite{ucf101}), satellite images (EuroSAT~\cite{eurosat}), and general objects (Caltech101~\cite{caltech101}).

While the main experiments are conducted using CLIP with a ViT-B-16 and ViT-L-14 backbone \cite{dosovitskiy2020image}, we later also explore additional architectures, including a ResNet-50 (RN50) and a ViT-H-14 model, all pretrained by OpenAI. Furthermore, we consider EVA-02-B-16 and EVA02-L-14 from \cite{sun2023eva}. Our evaluation metric is based on the error rate.

\subsubsection{Results}
\input{tables/prompts_detailed}
\input{figures/ablation_num_augmentations}
The results of the diverse prompt-based methods are illustrated in \Cref{tab:prompts_detailed}. Here, Source denotes employing zero-shot classification with different prompt strategies: utilizing the single prompt "\textit{a photo of a \{classname\}.}", an ensemble of hand-crafted prompt templates following \cite{radford2021learning}, the CuPL \cite{pratt2023does} prompts generated by an LLM, and a combination of both ensemble and CuPL prompts referred to as "All Prompts".

As shown in \Cref{tab:prompts_detailed}, all methods substantially improve on the single prompt baseline. While the LLM generated prompts of CuPL outperform the hand-crafted ensemble on five out of seven ImageNet variations for both architectures, better results can be achieved by leveraging all prompts. This even outperforms the optimization based approach TPT on four out of seven IN variations, while requiring only a fraction of its computational effort, i.e., one image forward versus 64 image forwards, 2 text forwards, and one backward. However, the best results are achieved by our VTE approach, which significantly outperforms all other baselines. Although this comes at the cost of an increased computational complexity compared to the hand-crafted approaches, VTE is still faster than TPT during inference due to not requiring any  backwards or text forwards through the respective encoder. We also find, that the performance of VTE can be further improved by employing a better prompt list. For ViT-B-16, for example, using All Prompts decreases the average error from $38.6\%$ to $38.0\%$.

In \cref{fig:num_augmentations}, we study the performance of VTE for different numbers of augmentations during test-time, employing the ensemble prompt and a ViT-B-16. Only applying 32 augmentations results in a mere $0.2\%$ increase in error rate compared to using 64 augmentations. Moreover, even with just 16 augmentations, there is still a notable improvement of $0.6\%$ in terms of average error rate compared to zero-shot classification with an ensemble prompt. 

\section{Updating Model Parameters with Test-Time Adaptation}
While the focus of the previous section has been on leveraging prompts and vision-text-space ensembles, in this section we want to put emphasis on a surprisingly underexplored topic, namely leveraging existing TTA methods for adapting vision-language models. The idea is straightforward, the text encoder of a CLIP model is frozen, allowing to precompute the text embeddings $\{\bm{t}_k\}_{k=1}^K$, ensuring efficiency during prediction. Given an image embedding $\bm{z}_t$, the cosine similarity can be computed $s_t = \mathrm{sim}(\bm{t}_k, \bm{z}_t)$. Treating the cosine similarities as the network's logits, the output probabilites can be received through \cref{eq:softmax_probs}. In this way we can treat any CLIP model as a common image classifier, enabling the application of any existing TTA method for image classification. Note that it is also possible to update the text encoder's parameters, but for now, we limit our analysis to only updating the parameters of the image encoder. In the following experiments, a batch size of 64 test samples per time step $t$ is employed.

\subsection{Test-Time Normalization for CLIP}
First, we investigate the performance of BN--1, a common procedure in TTA, which recalculates the batch normalization (BN) statistics using the current test batch. While Schneider et al. \cite{schneider2020improving} showed that recalculating the batch normalization statistics during test-time can significantly reduce the error rate for models pretrained on ImageNet, we investigate whether this is also the case for a CLIP model that was trained on millions of data samples covering a much broader data distribution. In \Cref{fig:bn_comparison} the zero-shot performance (source) and BN--1 performance is illustrated for CLIP with a RN50 and RN101 backbone using a single prompt. It can be clearly seen that, unlike for models pretrained on ImageNet, the average performance across the investigated datasets substantially decreases when applying BN--1. For RN50 the average error rate increases from 50.2\% to 74.1\% and for RN101 from 46.8\% to 71.0\%. This can be possibly attributed to much larger batch sizes and a much broader data distribution used during CLIP pretraining. A similar phenomenon is described in \cite{marsden2024universal}, where employing BN--1 for a regular ImageNet pretrained RN50 decreases the error rate on ImageNet-C from 82.0\% to 68.6\% in a continual TTA setting, but increases to 82.5\% in a mixed-domains TTA setting, where all corruptions of ImageNet-C are randomly suffled within the test sequence. Since BN--1 is employed by most TTA methods during adaptation, we conclude that RN backbones are not feasible. Instead we focus our following analysis on vision transformers that do not employ BN.
\input{figures/bn_comparison}

\subsection{Are Existing TTA Methods Beneficial for Vision-Language Foundation Models?}
\input{tables/tta_continual}
In this section, we take a deeper look into the performance of existing TTA methods applied to vision-language models, namely CLIP \cite{radford2021learning} and EVA-CLIP \cite{sun2023eva}. We evaluate influential and recent TTA methods: TENT \cite{wang2021tent}, ETA \cite{niu2022efficient}, SAR \cite{niu2023towards}, DeYO \cite{lee2024entropy}, CMF \cite{lee2024continual}, and ROID \cite{marsden2024universal} using the same adaptation setup and hyperparameters as proposed in the corresponding papers. We investigate ETA instead of EATA, since EATA requires access to samples from the source domain. In \Cref{tab:tta_continual} we report the error rate for CLIP with ViT-B-16 and ViT-L-14 backbones in the continual TTA setting \cite{wang2022continual}. We decide on the continual TTA setting, since this also shows how TTA methods cope with multiple distribution shifts. Later, in \Cref{sec:correlated}, we take a look into a more challenging scenario, namely dealing with temporally correlated test sequences.

All TTA methods improve on average upon the zero-shot performance for ViT-B-16 and ViT-L-14. ROID and CMF show a comparable performance and show the best performance for most datasets. ROID decreases the error rate on average by 1.9\% for ViT-B-16 and CMF by 1.5\% for ViT-L-14. It is noteworthy that even for the already strong source performance, both CMF and ROID are on-par or better than the zero-shot model for each considered dataset. Both CMF and ROID even outperform VTE and TPT despite their much higher compute cost. Comparing ROID and TPT, ROID is absolutely 1.6\% and 1.5\% better using a ViT-B-16 and ViT-L-14, respectively.

\paragraph{The importance of updating the vision encoder for certain distribution shifts}
\input{figures/umap}
Taking a closer look at the individual performances, interestingly, ROID and CMF show a relatively high improvement on ImageNet-C and EuroSAT. For a ViT-B-16 they roughly improve absolutely 8\% on ImageNet-C and 15\% on EuroSAT compared to the source baseline. Getting insights into this phenomenon, we illustrate the feature space of the ViT-B-16 backbone before and after adaptation (adapted with ROID) for EuroSAT and compare it to the dataset Pets, where no significant improvement is seen. The UMAP visualization is shown in \Cref{fig:umap}. Comparing the low-dimensional space of EuroSAT and Pets before adaptation, it can be clearly seen that the zero-shot model has a much better class separation for Pets than for EuroSAT. For EuroSAT there is significant class overlap, hence, updating the vision encoder can result in a much more discriminative feature space. This undermines the importance and opportunity of adapting the vision encoder for data distributions where the zero-shot model has limited class separation. This also shows the limitations of prompt-based methods which simply work with a fixed image feature space.

\paragraph{Test-time adaptation remains beneficial for large models}
\input{figures/model_comparison}
A natural question that arises is whether the improvement for TTA methods diminishes for bigger models with a better initial performance. Therefore, in \Cref{fig:model_comparison}, the error rate is illustrated for ViT-B-16 up to ViT-H-14. As one would expect, the performance gains through adaptation diminishes as the zero-shot performance improves. But, even for a ViT-H-14, all investigated TTA methods still improve upon the source performance with CMF taking the lead, reducing the error rate further by absolutely 1\%. Interestingly, in contrast to the CLIP models by OpenAI, for the investigated EVA-CLIP models not all TTA methods, namely TENT and DeYO, can improve upon the zero-shot model. 

\subsection{Test-Time Adaptation for Non-i.i.d. Data Streams}
\label{sec:correlated}
\input{tables/tta_correlated}
Since the previously investigated continual TTA setting might not apply to real-world online data streams, we additionally investigate a scenario with temporally correlated samples. In the correlated TTA setting the data of each domain is sorted by the class label rather than randomly shuffled, resulting in class-imbalanced batches. The results are reported in \Cref{tab:tta_correlated}. For the more challenging correlated TTA setting, in contrast to the continual setting, not all TTA methods are capable to improve upon the source performance. Only CMF and ROID show a stable adaptation. Due to the employed prior correction\footnote{In contrast to the original prior correction, we find that applying the prior correction in the output probability space instead of the logit space shows a more consistent performance for the investigated CLIP models.} proposed in \cite{marsden2024universal}, CMF and ROID perform even better than in the continual setting.

\subsection{Updating the Text Encoder}
Up to now, only the parameters or a subset of the parameters of the vision encoder were updated. Additionally updating the text encoder comes with a non-neglectable overhead. In this case, all text prompts have to be forwarded through the text encoder each update step. E.g., when using the common text prompt ensemble for ImageNet, this would require forwarding 80,000 text prompts each step and can quickly lead to an explosion in memory or compute requirement. Therefore, we restrict our ablation to using a single prompt for a ViT-B-16 backbone. Considering TENT, additionally updating the text encoder, decreases the performance on average by 1.8\%. ROID improves on average by 0.2\%, but compared to the ROID variant that employs the text ensemble, updating the text encoder with a single prompt is still 0.9\% behind. Given these outcomes, we can conclude that updating the text encoder in addition to updating the vision encoder is not beneficial.

\subsection{Limitations}
A limitation of applying existing TTA methods to vision-language models is that they often require a batch of test data at each time step $t$ for effective parameter updates. However, as discussed in \cite{marsden2024universal}, this is only partially true. Networks that do not employ BN layers, such as VisionTransformer \cite{dosovitskiy2020image}, allow to recover the batch TTA setting by simply accumulating the gradients of the last $b$ test samples before updating the model. This comes with no computational overhead and even significantly reduces the memory requirement.

\section{Conclusion}
In this work, we explored the task of adapting vision-language models at test-time to accommodate distribution shifts. Our investigation led us through a comprehensive analysis of both prompt-based approaches and existing test-time adaptation (TTA) methods applied to vision-language models, focusing particularly on CLIP and its variants. Our introduced vision-text-space ensemble shows to be the better option when compared to TPT. Our exploration of existing TTA methods revealed their potential for enhancing the robustness of vision-language models. Methods like ROID and CMF showcased impressive performance improvements across various datasets and model architectures.

\newpage
%
%
\bibliographystyle{splncs04}
\bibliography{main}
\end{document}

%% file: figures/overview.tex
\begin{figure}[t]
\centering
\scalebox{0.9}{
\def\svgwidth{250pt}
\graphicspath{{figures/}}
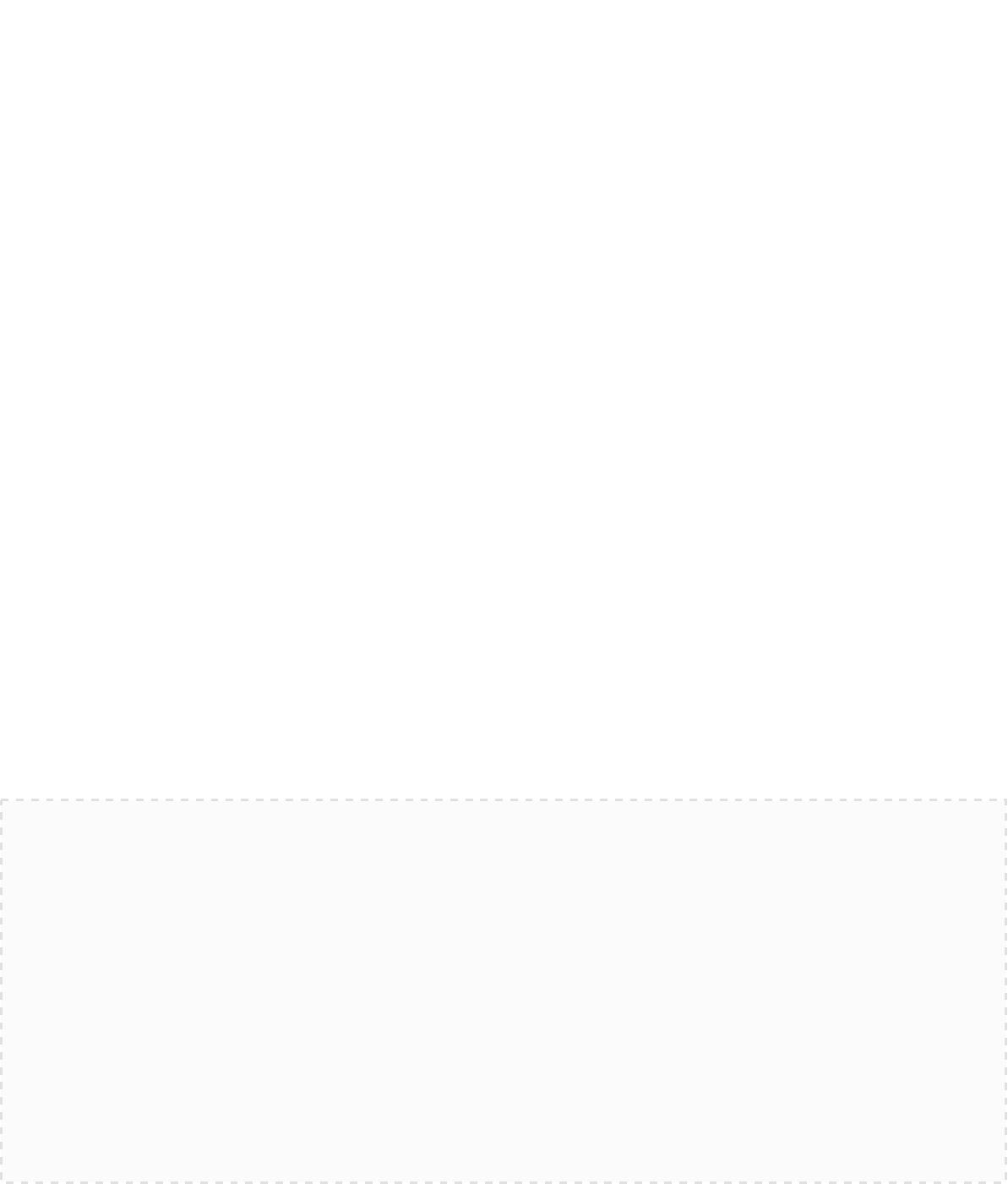
}
\caption{Overview of the proposed VTE approach and the application of existing TTA methods for VLMs. Before inference, an average text representation $\bar{\bm{t}}_k$ for each of the $K$ classes is extracted by mapping a list of prompts into the text embedding space. During inference, VTE uses test-time augmentation and entropy based filtering. In the case of applying TTA methods, only the parameters of the vision encoder are updated.}
\label{fig:settings}
\end{figure}

%% file: figures/CLIP_TTA.pdf_tex
\begingroup%
  \makeatletter%
  \providecommand\color[2][]{%
    \errmessage{(Inkscape) Color is used for the text in Inkscape, but the package 'color.sty' is not loaded}%
    \renewcommand\color[2][]{}%
  }%
  \providecommand\transparent[1]{%
    \errmessage{(Inkscape) Transparency is used (non-zero) for the text in Inkscape, but the package 'transparent.sty' is not loaded}%
    \renewcommand\transparent[1]{}%
  }%
  \providecommand\rotatebox[2]{#2}%
  \newcommand*\fsize{\dimexpr\f@size pt\relax}%
  \newcommand*\lineheight[1]{\fontsize{\fsize}{#1\fsize}\selectfont}%
  \ifx\svgwidth\undefined%
    \setlength{\unitlength}{572.40613201bp}%
    \ifx\svgscale\undefined%
      \relax%
    \else%
      \setlength{\unitlength}{\unitlength * \real{\svgscale}}%
    \fi%
  \else%
    \setlength{\unitlength}{\svgwidth}%
  \fi%
  \global\let\svgwidth\undefined%
  \global\let\svgscale\undefined%
  \makeatother%
  \begin{picture}(1,1.17442574)%
    \lineheight{1}%
    \setlength\tabcolsep{0pt}%
    \put(0,0){\includegraphics[width=\unitlength,page=1]{CLIP_TTA.pdf}}%
    \put(0.0731078,0.07825837){\color[rgb]{0,0,0}\makebox(0,0)[lt]{\lineheight{1.25}\smash{\begin{tabular}[t]{l}\small{frozen}\\\small{trainable}\end{tabular}}}}%
    \put(0,0){\includegraphics[width=\unitlength,page=2]{CLIP_TTA.pdf}}%
    \put(0.36296981,0.96653432){\color[rgb]{0,0,0}\makebox(0,0)[lt]{\lineheight{1.25}\smash{\begin{tabular}[t]{l}\small{text}\\\small{encoder}\end{tabular}}}}%
    \put(0.03999404,1.03656386){\color[rgb]{0,0,0}\makebox(0,0)[lt]{\lineheight{1.25}\smash{\begin{tabular}[t]{l}\small{cat}\end{tabular}}}}%
    \put(0.16792097,0.98950555){\color[rgb]{0,0,0}\makebox(0,0)[lt]{\lineheight{1.25}\smash{\begin{tabular}[t]{l}\small{prompt}\\\small{template}\\\small{list}\end{tabular}}}}%
    \put(0.04040737,0.98326174){\color[rgb]{0,0,0}\makebox(0,0)[lt]{\lineheight{1.25}\smash{\begin{tabular}[t]{l}\small{dog}\end{tabular}}}}%
    \put(0.04010092,0.84387612){\color[rgb]{0,0,0}\makebox(0,0)[lt]{\lineheight{1.25}\smash{\begin{tabular}[t]{l}\small{bird}\end{tabular}}}}%
    \put(0,0){\includegraphics[width=\unitlength,page=3]{CLIP_TTA.pdf}}%
    \put(0.02418147,1.13034255){\color[rgb]{0,0,0}\makebox(0,0)[lt]{\lineheight{1.25}\smash{\begin{tabular}[t]{l}\text{Before inference}\end{tabular}}}}%
    \put(0,0){\includegraphics[width=\unitlength,page=4]{CLIP_TTA.pdf}}%
    \put(0.57761101,1.03718327){\color[rgb]{0,0,0}\makebox(0,0)[lt]{\lineheight{1.25}\smash{\begin{tabular}[t]{l}\small{$\bm{t}_{11}\dots\,\bm{t}_{1J}$}\end{tabular}}}}%
    \put(0.78514683,0.96472796){\color[rgb]{0,0,0}\makebox(0,0)[lt]{\lineheight{1.25}\smash{\begin{tabular}[t]{l}\small{avg.}\end{tabular}}}}%
    \put(0.57801945,0.98388102){\color[rgb]{0,0,0}\makebox(0,0)[lt]{\lineheight{1.25}\smash{\begin{tabular}[t]{l}\small{$\bm{t}_{21}\dots\,\bm{t}_{2J}$}\end{tabular}}}}%
    \put(0.57772282,0.84449539){\color[rgb]{0,0,0}\makebox(0,0)[lt]{\lineheight{1.25}\smash{\begin{tabular}[t]{l}\small{$\bm{t}_{K1}\dots\,\bm{t}_{KJ}$}\end{tabular}}}}%
    \put(0,0){\includegraphics[width=\unitlength,page=5]{CLIP_TTA.pdf}}%
    \put(0.8711487,1.03711977){\color[rgb]{0,0,0}\makebox(0,0)[lt]{\lineheight{1.25}\smash{\begin{tabular}[t]{l}\small{$\overline{\bm{t}}_1$}\end{tabular}}}}%
    \put(0.87155928,0.98381753){\color[rgb]{0,0,0}\makebox(0,0)[lt]{\lineheight{1.25}\smash{\begin{tabular}[t]{l}\small{$\overline{\bm{t}}_2$}\end{tabular}}}}%
    \put(0.87125134,0.84443191){\color[rgb]{0,0,0}\makebox(0,0)[lt]{\lineheight{1.25}\smash{\begin{tabular}[t]{l}\small{$\overline{\bm{t}}_K$}\end{tabular}}}}%
    \put(0,0){\includegraphics[width=\unitlength,page=6]{CLIP_TTA.pdf}}%
    \put(0.36282267,0.56961173){\color[rgb]{0,0,0}\makebox(0,0)[lt]{\lineheight{1.25}\smash{\begin{tabular}[t]{l}\small{image}\\\small{encoder}\end{tabular}}}}%
    \put(0.2285241,0.63964126){\color[rgb]{0,0,0}\makebox(0,0)[lt]{\lineheight{1.25}\smash{\begin{tabular}[t]{l}\small{$\bm{x}_{t1}$}\end{tabular}}}}%
    \put(0.09846899,0.5452593){\color[rgb]{0,0,0}\makebox(0,0)[lt]{\lineheight{1.25}\smash{\begin{tabular}[t]{l}\small{Aug}\end{tabular}}}}%
    \put(0.22893744,0.58633915){\color[rgb]{0,0,0}\makebox(0,0)[lt]{\lineheight{1.25}\smash{\begin{tabular}[t]{l}\small{$\bm{x}_{t2}$}\end{tabular}}}}%
    \put(0.22863099,0.44695353){\color[rgb]{0,0,0}\makebox(0,0)[lt]{\lineheight{1.25}\smash{\begin{tabular}[t]{l}\small{$\bm{x}_{tB}$}\end{tabular}}}}%
    \put(0,0){\includegraphics[width=\unitlength,page=7]{CLIP_TTA.pdf}}%
    \put(0.02403431,0.73342007){\color[rgb]{0,0,0}\makebox(0,0)[lt]{\lineheight{1.25}\smash{\begin{tabular}[t]{l}\text{VTE during inference}\end{tabular}}}}%
    \put(0,0){\includegraphics[width=\unitlength,page=8]{CLIP_TTA.pdf}}%
    \put(0.55982838,0.64003498){\color[rgb]{0,0,0}\makebox(0,0)[lt]{\lineheight{1.25}\smash{\begin{tabular}[t]{l}\small{$\bm{z}_{t1}$}\end{tabular}}}}%
    \put(0.56027149,0.58673294){\color[rgb]{0,0,0}\makebox(0,0)[lt]{\lineheight{1.25}\smash{\begin{tabular}[t]{l}\small{$\bm{z}_{t2}$}\end{tabular}}}}%
    \put(0.55996477,0.44734733){\color[rgb]{0,0,0}\makebox(0,0)[lt]{\lineheight{1.25}\smash{\begin{tabular}[t]{l}\small{$\bm{z}_{tB}$}\end{tabular}}}}%
    \put(0,0){\includegraphics[width=\unitlength,page=9]{CLIP_TTA.pdf}}%
    \put(0.7110785,0.56949415){\color[rgb]{0,0,0}\makebox(0,0)[lt]{\lineheight{1.25}\smash{\begin{tabular}[t]{l}\small{avg.}\end{tabular}}}}%
    \put(0,0){\includegraphics[width=\unitlength,page=10]{CLIP_TTA.pdf}}%
    \put(0.36249513,0.17358592){\color[rgb]{0,0,0}\makebox(0,0)[lt]{\lineheight{1.25}\smash{\begin{tabular}[t]{l}\small{image}\\\small{encoder}\end{tabular}}}}%
    \put(0.84163071,0.14827319){\color[rgb]{0,0,0}\makebox(0,0)[lt]{\lineheight{1.25}\smash{\begin{tabular}[t]{l}\small{$\mathcal{L}(\bm{p}_t)$}\end{tabular}}}}%
    \put(0,0){\includegraphics[width=\unitlength,page=11]{CLIP_TTA.pdf}}%
    \put(0.02370673,0.3373945){\color[rgb]{0,0,0}\makebox(0,0)[lt]{\lineheight{1.25}\smash{\begin{tabular}[t]{l}\text{TTA methods during inference}\end{tabular}}}}%
    \put(0,0){\includegraphics[width=\unitlength,page=12]{CLIP_TTA.pdf}}%
    \put(0.64008924,0.2704405){\color[rgb]{0,0,0}\makebox(0,0)[lt]{\lineheight{1.25}\smash{\begin{tabular}[t]{l}\small{$\overline{\bm{t}}_{1}\dots \overline{\bm{t}}_{K}$}\end{tabular}}}}%
    \put(0.61473628,0.16796758){\color[rgb]{0,0,0}\makebox(0,0)[lt]{\lineheight{1.25}\smash{\begin{tabular}[t]{l}\small{cosine sim.}\\\small{+ softmax}\end{tabular}}}}%
    \put(0,0){\includegraphics[width=\unitlength,page=13]{CLIP_TTA.pdf}}%
    \put(0.82656524,0.66657747){\color[rgb]{0,0,0}\makebox(0,0)[lt]{\lineheight{1.25}\smash{\begin{tabular}[t]{l}\small{$\overline{\bm{t}}_{1}\dots \overline{\bm{t}}_{K}$}\end{tabular}}}}%
    \put(0.80102,0.56410469){\color[rgb]{0,0,0}\makebox(0,0)[lt]{\lineheight{1.25}\smash{\begin{tabular}[t]{l}\small{cosine sim.}\\\small{+ softmax}\end{tabular}}}}%
    \put(0,0){\includegraphics[width=\unitlength,page=14]{CLIP_TTA.pdf}}%
    \put(0.23911095,0.15535728){\color[rgb]{0,0,0}\makebox(0,0)[lt]{\lineheight{1.25}\smash{\begin{tabular}[t]{l}\small{$\bm{x}_t$}\end{tabular}}}}%
    \put(0,0){\includegraphics[width=\unitlength,page=15]{CLIP_TTA.pdf}}%
    \put(0.02145639,0.54890508){\color[rgb]{0,0,0}\makebox(0,0)[lt]{\lineheight{1.25}\smash{\begin{tabular}[t]{l}\small{$\bm{x}_t$}\end{tabular}}}}%
    \put(0.53406554,0.17180432){\color[rgb]{0,0,0}\makebox(0,0)[lt]{\lineheight{1.25}\smash{\begin{tabular}[t]{l}\small{$\bm{z}_t$}\end{tabular}}}}%
    \put(0,0){\includegraphics[width=\unitlength,page=16]{CLIP_TTA.pdf}}%
  \end{picture}%
\endgroup%

%% file: tables/prompts_detailed.tex
\begin{table}[t]
\renewcommand{\arraystretch}{1.2}
\centering
\caption{Online classification error rate~(\%) for CLIP with a ViT-B-16 and ViT-L-14 backbone pretrained by OpenAI. The models comprise 149.62 million parameters with 41.09 billion FLOPS and 427.62 million parameters with 175.33 billion FLOPS, respectively.}
\vskip -0.1in
\scalebox{0.62}{
\tabcolsep2pt
\begin{tabular}{l|l|l|ccccccccccccccccc|c}\hline
    & Method & Prompt & \rotatebox[origin=c]{70}{ImageNet} & \rotatebox[origin=c]{70}{ImageNet-C} & \rotatebox[origin=c]{70}{ImageNet-A} & \rotatebox[origin=c]{70}{ImageNet-V2} & \rotatebox[origin=c]{70}{ImageNet-R} & \rotatebox[origin=c]{70}{ImageNet-S} & \rotatebox[origin=c]{70}{ImageNet-D109} & \rotatebox[origin=c]{70}{Flower102} & \rotatebox[origin=c]{70}{DTD} & \rotatebox[origin=c]{70}{Pets} & \rotatebox[origin=c]{70}{Cars} & \rotatebox[origin=c]{70}{UCF101} & \rotatebox[origin=c]{70}{Caltech101} & \rotatebox[origin=c]{70}{Food101} & \rotatebox[origin=c]{70}{SUN397} & \rotatebox[origin=c]{70}{Aircraft} & \rotatebox[origin=c]{70}{EuroSAT} & Avg. \\
    \hline\hline
\multirow{7}{*}{\rotatebox[origin=c]{90}{ViT-B-16}} & \multirow{4}{*}{Source} & Single & 33.3 & 75.5 & 52.3 & 39.2 & 26.0 & 53.9 & 29.5 & 32.6 & 55.4 & \textbf{11.8} & 34.8 & 34.9 & 7.1 & 16.2 & 37.4 & 76.2 & 58.5 & 39.7\\ 
&  & Ensemble & 31.7 & 73.8 & 49.9 & 38.1 & 22.5 & 51.7 & 27.5 & 34.2 & 54.6 & \textbf{11.8} & 33.6 & 32.6 & 7.0 & 15.5 & 34.6 & 76.5 & \textbf{51.8} & 38.1\\ 
&  & CuPL & 30.4 & 73.3 & 49.3 & 36.7 & 22.9 & 51.0 & 28.0 & - & - & - & - & - & - & - & - & - & - & -\\ 
&  & All Prompts & 30.3 & \textbf{73.0} & 49.0 & 36.9 & 22.0 & 50.6 & 27.2 & - & - & - & - & - & - & - & - & - & - & -\\ 
\cline{2-21}
& TPT & a photo of a & 31.0 & 75.1 & 45.7 & 36.5 & 23.0 & 52.2 & 26.8 & \textbf{30.9} & \textbf{52.7} & 12.8 & 34.0 & \textbf{32.5} & \textbf{6.3} & \textbf{15.2} & 34.6 & 76.7 & 57.2 & 37.8\\ 
& VTE & Ensemble & 29.6 & 74.4 & 37.3 & 34.9 & \textbf{19.6} & 49.8 & 24.6 & 34.5 & 52.7 & 13.0 & \textbf{31.0} & 33.0 & 6.7 & 16.6 & \textbf{33.5} & \textbf{75.9} & 52.4 & \textbf{36.4}\\ 
& VTE & All Prompts & \textbf{28.3} & 73.6 & \textbf{36.7} & \textbf{34.1} & 19.7 & \textbf{49.0} & \textbf{24.5} & - & - & - & - & - & - & - & - & - & - & -\\
\hline
\hline
\multirow{7}{*}{\rotatebox[origin=c]{90}{ViT-L-14}} & \multirow{4}{*}{Source} & Single & 26.5 & 60.5 & 31.3 & 32.1 & 14.6 & 42.1 & 24.2 & 24.1 & 47.4 & 6.8 & 23.2 & 27.3 & 5.2 & 11.4 & 32.5 & 69.7 & 44.7 & 30.8\\ 
&  & Ensemble & 24.5 & 58.6 & 29.3 & 30.1 & 12.2 & 40.4 & 22.4 & 24.4 & 43.3 & 7.0 & 22.1 & 25.0 & 5.5 & \textbf{10.8} & 30.9 & 68.1 & \textbf{39.3} & 29.1\\ 
&  & CuPL & 23.4 & 57.7 & 28.2 & 29.2 & 12.3 & 40.0 & 22.5 & - & - & - & - & - & - & - & - & - & - & -\\ 
&  & All Prompts & 23.5 & \textbf{57.5} & 28.2 & 29.0 & 11.8 & 39.7 & 22.1 & - & - & - & - & - & - & - & - & - & - & -\\ 
\cline{2-21}
& TPT & a photo of a & 24.5 & 59.0 & 25.2 & 29.9 & 12.1 & 40.2 & 22.0 & \textbf{23.5} & 46.1 & \textbf{6.4} & 22.3 & 25.5 & 4.4 & 10.9 & 29.8 & 68.7 & 48.0 & 29.3\\ 
& VTE & Ensemble & 23.0 & 59.5 & 20.4 & 28.4 & 10.3 & 38.9 & 20.5 & 26.2 & \textbf{41.9} & 7.1 & \textbf{21.6} & \textbf{24.6} & \textbf{4.2} & 11.7 & \textbf{29.3} & \textbf{66.1} & 46.4 & \textbf{28.2}\\ 
& VTE & All Prompts & \textbf{22.3} & 58.8 & \textbf{19.7} & \textbf{27.5} & \textbf{9.8} & \textbf{38.3} & \textbf{20.2} & - & - & - & - & - & - & - & - & - & - & -\\
    \hline
\end{tabular}}
\label{tab:prompts_detailed}
\end{table}

%% file: figures/ablation_num_augmentations.tex
\begin{figure}[t]
\centering
\includegraphics[scale=0.68]{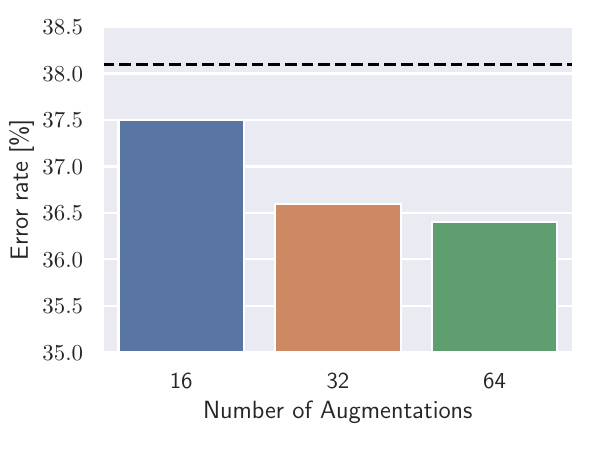}
\vspace{-4mm}
\caption{Average error rate of VTE with a ViT-B-16 backbone across all 17 datasets when using different numbers of augmentations during test-time. The dashed line indicates the performance of zero-shot CLIP with Ensemble prompts.}
\label{fig:num_augmentations}
\end{figure}

%% file: figures/bn_comparison.tex
\begin{figure}[t]
\centering
\includegraphics[scale=0.68]{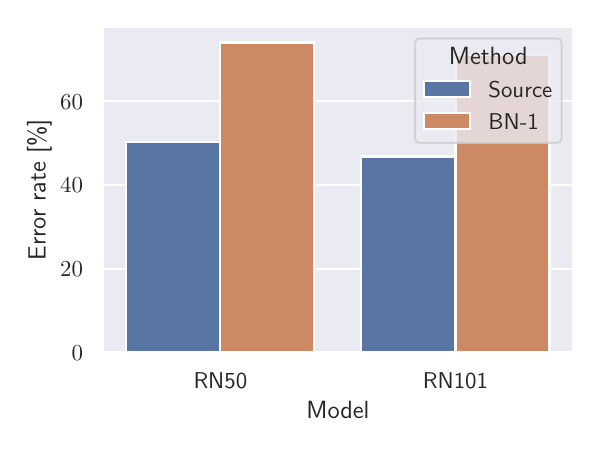}
\vspace{-4mm}
\caption{Average error rate for CLIP with a RN50 and RN101 backbone for both source and BN--1. As illustrated, the error rate drastically increases when the normalization statistics are recalculated during test-time.}
\label{fig:bn_comparison}
\end{figure}

%% file: tables/tta_continual.tex
\begin{table}[t]
\renewcommand{\arraystretch}{1.2}
\centering
\caption{Online classification error rate~(\%) for CLIP with a ViT-B-16 and ViT-L-14 backbone pretrained by OpenAI in a continual TTA setting. The models comprise 149.62 million parameters with 41.09 billion FLOPS and 427.62 million parameters with 175.33 billion FLOPS, respectively.}
\vskip -0.1in
\scalebox{0.63}{
\tabcolsep2pt
\begin{tabular}{l|l|l|ccccccccccccccccc|c}\hline
    & Method & Prompt & \rotatebox[origin=c]{70}{ImageNet} & \rotatebox[origin=c]{70}{ImageNet-C} & \rotatebox[origin=c]{70}{ImageNet-A} & \rotatebox[origin=c]{70}{ImageNet-V2} & \rotatebox[origin=c]{70}{ImageNet-R} & \rotatebox[origin=c]{70}{ImageNet-S} & \rotatebox[origin=c]{70}{ImageNet-D109} & \rotatebox[origin=c]{70}{Flower102} & \rotatebox[origin=c]{70}{DTD} & \rotatebox[origin=c]{70}{Pets} & \rotatebox[origin=c]{70}{Cars} & \rotatebox[origin=c]{70}{UCF101} & \rotatebox[origin=c]{70}{Caltech101} & \rotatebox[origin=c]{70}{Food101} & \rotatebox[origin=c]{70}{SUN397} & \rotatebox[origin=c]{70}{Aircraft} & \rotatebox[origin=c]{70}{EuroSAT} & Avg. \\
    \hline\hline
\multirow{7}{*}{\rotatebox[origin=c]{90}{ViT-B-16}} & Source & Ensemble & 31.7 & 73.8 & 49.9 & \textbf{38.1} & 22.5 & 51.7 & 27.5 & 34.2 & 54.6 & 11.8 & 33.6 & 32.6 & 7.0 & \textbf{15.5} & 34.6 & 76.5 & 51.8 & 38.1\\ 
\cline{2-21}
& TENT & Ensemble & \textbf{31.6} & 75.4 & 49.6 & \textbf{38.1} & 21.6 & 51.6 & 27.2 & 34.2 & 54.3 & 11.6 & 33.6 & 32.3 & 6.9 & 15.8 & 34.4 & 76.5 & 43.3 & 37.5\\ 
& ETA & Ensemble & 32.1 & 69.1 & 49.3 & 38.3 & 21.7 & 50.9 & 26.9 & 34.0 & 54.4 & \textbf{11.3} & 33.5 & 32.1 & 7.0 & 16.0 & 33.9 & 76.2 & 50.7 & 37.5\\ 
& SAR & Ensemble & 32.0 & 70.0 & \textbf{49.2} & 38.3 & 21.8 & 51.5 & 27.0 & 33.9 & 54.4 & 11.5 & 34.1 & 32.5 & 7.0 & 15.7 & 34.8 & 76.4 & 44.7 & 37.3\\ 
& DeYO & Ensemble & 32.4 & 71.3 & 48.9 & 38.4 & 21.7 & 51.5 & 27.0 & 34.2 & 54.3 & \textbf{11.3} & 34.1 & 32.1 & 7.2 & 16.0 & 35.0 & 76.2 & 50.1 & 37.7\\ 
& CMF & Ensemble & 31.9 & 66.1 & 49.6 & 38.3 & \textbf{20.9} & \textbf{50.4} & \textbf{25.7} & 33.8 & 54.4 & 11.5 & 33.8 & 32.0 & \textbf{6.9} & 15.7 & 33.6 & 76.3 & 36.5 & 36.3\\ 
& ROID & Ensemble & 31.7 & \textbf{65.7} & 49.3 & 38.2 & 21.1 & 50.9 & 26.3 & \textbf{33.6} & \textbf{54.2} & 11.4 & \textbf{33.4} & \textbf{31.9} & \textbf{6.9} & 15.8 & \textbf{33.4} & \textbf{76.1} & \textbf{36.3} & \textbf{36.2} \\
\hline\hline
\multirow{7}{*}{\rotatebox[origin=c]{90}{ViT-L-14}} & Source & Ensemble & 24.5 & 58.6 & 29.3 & 30.1 & 12.2 & 40.4 & 22.4 & 24.4 & 43.3 & 7.0 & 22.1 & 25.0 & \textbf{5.5} & 10.8 & 30.9 & 68.1 & 39.3 & 29.1\\ 
\cline{2-21}
& TENT & Ensemble & 24.6 & 56.1 & 29.3 & 30.3 & 12.1 & 40.1 & 22.1 & 24.4 & 43.2 & 6.9 & 22.2 & 24.9 & 5.6 & 10.8 & 30.8 & 68.0 & 36.5 & 28.7\\ 
& ETA & Ensemble & 24.6 & 53.8 & 28.9 & 30.4 & 11.9 & 39.6 & 21.7 & 24.3 & 43.1 & 7.0 & \textbf{21.7} & 24.7 & 5.6 & 10.8 & 30.5 & 67.8 & 39.3 & 28.6\\ 
& SAR & Ensemble & 24.6 & 54.9 & 28.9 & 30.3 & 11.8 & 39.8 & 21.9 & 24.3 & 43.2 & 6.9 & 22.1 & 24.9 & 5.6 & \textbf{10.7} & 30.4 & 68.0 & 36.2 & 28.5\\ 
& DeYO & Ensemble & 24.6 & 54.3 & 28.6 & 30.5 & 11.6 & 39.6 & 21.3 & 24.3 & 43.1 & 6.8 & 21.8 & 24.9 & 5.7 & 10.9 & 30.7 & 67.9 & 38.1 & 28.5\\ 
& CMF & Ensemble & \textbf{24.2} & \textbf{50.6} & \textbf{28.2} & \textbf{30.0} & \textbf{11.1} & \textbf{38.6} & \textbf{20.4} & 24.1 & \textbf{43.0} & \textbf{6.6} & 21.9 & 24.7 & 5.6 & \textbf{10.7} & \textbf{30.1} & \textbf{67.4} & \textbf{32.2} & \textbf{27.6}\\ 
& ROID & Ensemble & 24.3 & 51.4 & 28.4 & \textbf{30.0} & 11.6 & 39.3 & 21.5 & \textbf{23.9} & 43.3 & \textbf{6.6} & 21.8 & \textbf{24.7} & 5.6 & \textbf{10.7} & \textbf{30.1} & 67.6 & 32.3 & 27.8\\
    \hline
\end{tabular}}
\label{tab:tta_continual}
\end{table}

%% file: figures/umap.tex
\begin{figure}[t]
\centering
\begin{tabular}{cc}
        \includegraphics[scale=0.35]{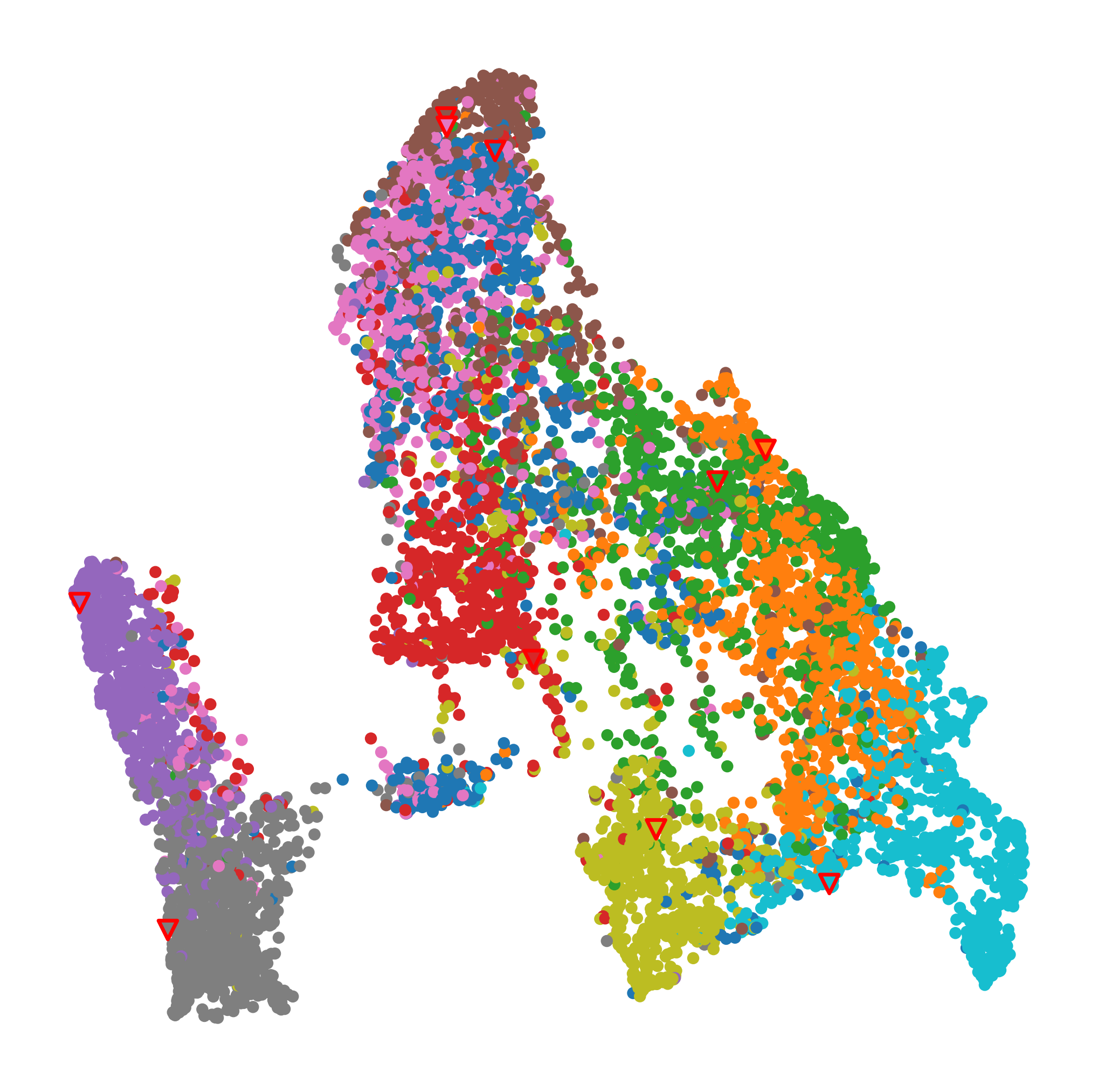} & \includegraphics[scale=0.35]{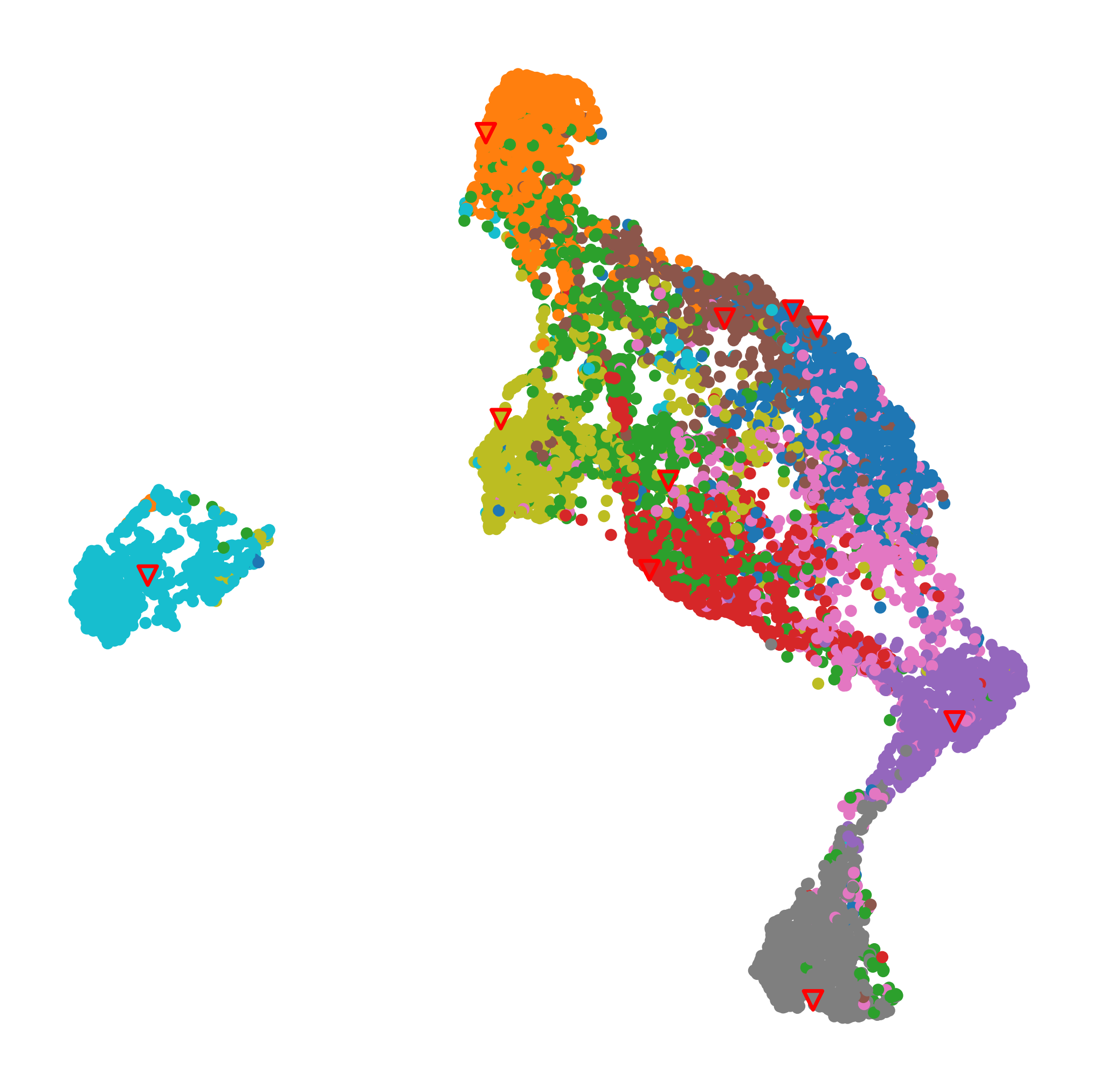} \\
        \includegraphics[scale=0.35]{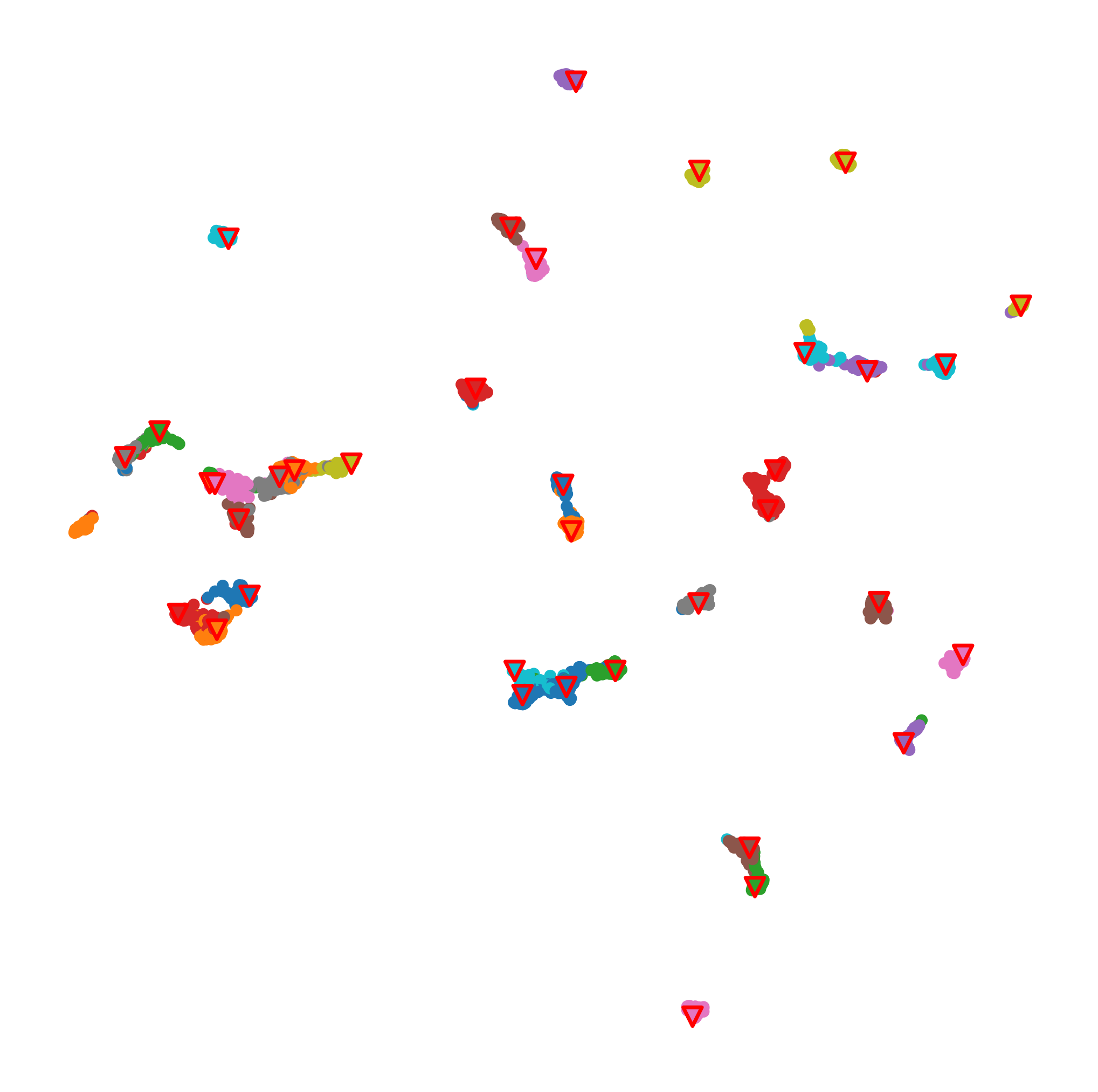} & \includegraphics[scale=0.35]{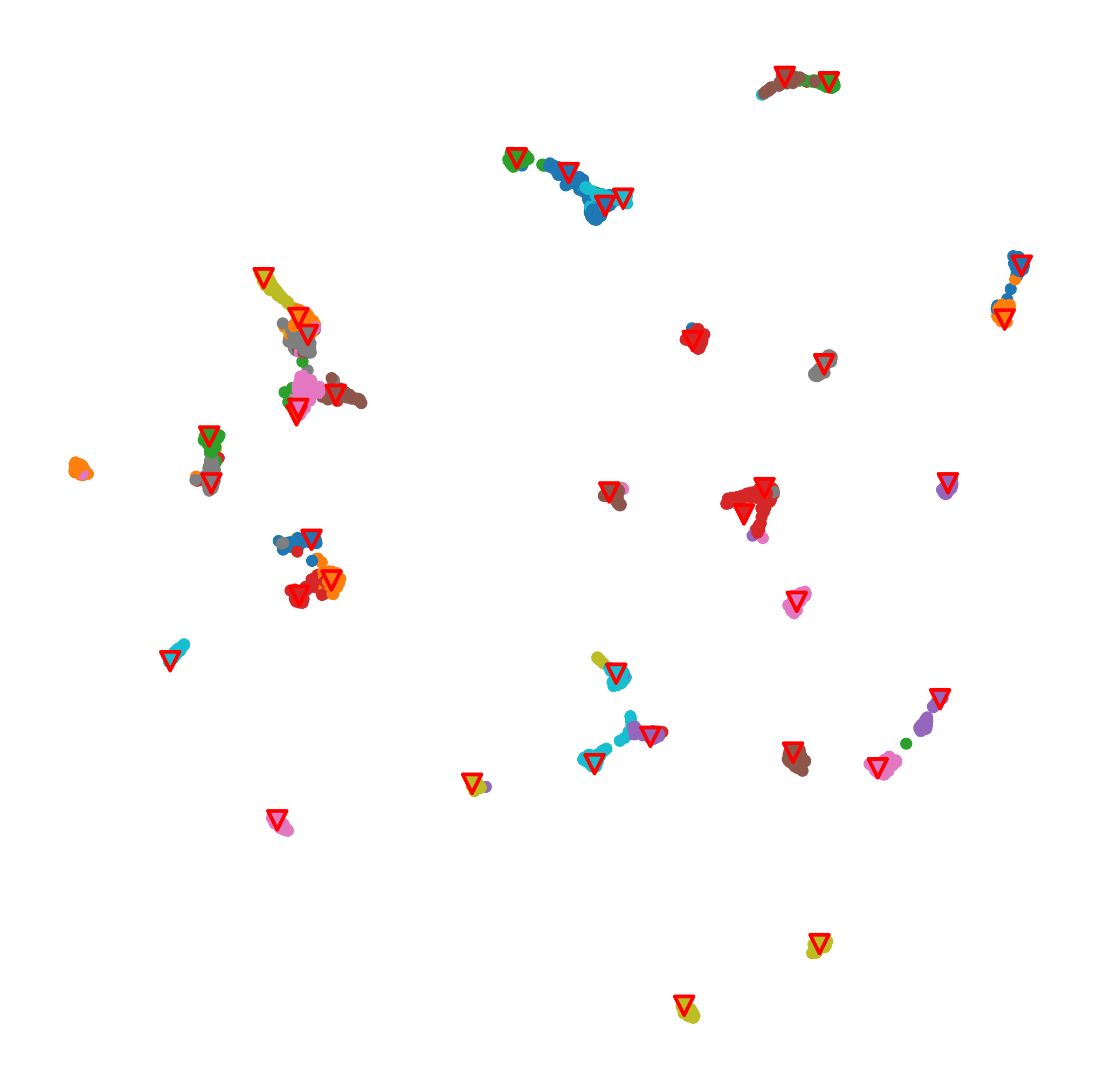} \\
    \end{tabular}
\vspace{-4mm}
\caption{UMAP visualization for EuroSAT (top) and Pets (bottom) before (left) and after adaptation (right). To better align the text and image embeddings, we use a projection proposed in \cite{hu2024reclip} before applying UMAP. The triangles illustrate the corresponding text ensemble embeddings.}
\label{fig:umap}
\end{figure}

%% file: figures/model_comparison.tex
\begin{figure}[t]
\centering
\includegraphics[scale=0.68]{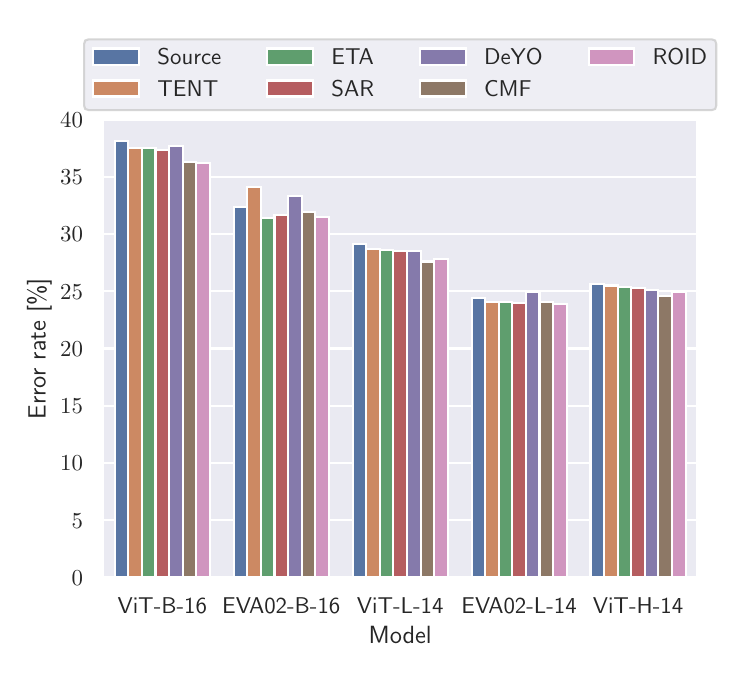}
\caption{Comparison of different models sorted according to their number of parameters from low (left) to high (right). The average error rate across all datasets is reported.}
\label{fig:model_comparison}
\end{figure}

%% file: tables/tta_correlated.tex
\begin{table}[t]
\renewcommand{\arraystretch}{1.2}
\centering
\caption{Online classification error rate~(\%) for CLIP with a ViT-B-16 and ViT-L-14 backbone pretrained by OpenAI in a correlated TTA setting. The models comprise 149.62 million parameters with 41.09 billion FLOPS and 427.62 million parameters with 175.33 billion FLOPS, respectively.}
\vskip -0.1in
\scalebox{0.63}{
\tabcolsep2pt
\begin{tabular}{l|l|l|ccccccccccccccccc|c}\hline
    & Method & Prompt & \rotatebox[origin=c]{70}{ImageNet} & \rotatebox[origin=c]{70}{ImageNet-C} & \rotatebox[origin=c]{70}{ImageNet-A} & \rotatebox[origin=c]{70}{ImageNet-V2} & \rotatebox[origin=c]{70}{ImageNet-R} & \rotatebox[origin=c]{70}{ImageNet-S} & \rotatebox[origin=c]{70}{ImageNet-D109} & \rotatebox[origin=c]{70}{Flower102} & \rotatebox[origin=c]{70}{DTD} & \rotatebox[origin=c]{70}{Pets} & \rotatebox[origin=c]{70}{Cars} & \rotatebox[origin=c]{70}{UCF101} & \rotatebox[origin=c]{70}{Caltech101} & \rotatebox[origin=c]{70}{Food101} & \rotatebox[origin=c]{70}{SUN397} & \rotatebox[origin=c]{70}{Aircraft} & \rotatebox[origin=c]{70}{EuroSAT} & Avg. \\
    \hline\hline
\multirow{7}{*}{\rotatebox[origin=c]{90}{ViT-B-16}} & Source & Ensemble & 31.7 & 73.8 & 49.9 & 38.1 & 22.5 & 51.7 & 27.5 & 34.2 & 54.6 & 11.8 & 33.6 & 32.6 & 7.0 & 15.5 & 34.6 & 76.5 & 51.8 & 38.1\\ 
\cline{2-21}
& TENT & Ensemble & 31.6 & 93.9 & 49.6 & 38.0 & 21.8 & 51.8 & 27.3 & 34.1 & 54.4 & 11.5 & 33.8 & 32.6 & 6.9 & 15.9 & 34.6 & 76.3 & 41.1 & 38.5\\
& ETA & Ensemble & 33.7 & 88.4 & 51.4 & 38.1 & 23.0 & 53.6 & 30.9 & 34.1 & 54.3 & 11.5 & 33.8 & 32.7 & 7.0 & 20.5 & 34.5 & 76.5 & 51.7 & 39.7\\
& SAR & Ensemble & 32.1 & 69.9 & 49.0 & 38.4 & 22.0 & 52.3 & 27.4 & 34.6 & 54.6 & 11.2 & 33.9 & 32.7 & 7.1 & 16.3 & 34.9 & 76.5 & 47.9 & 37.7\\
& DeYO & Ensemble & 32.7 & 99.7 & 49.6 & 38.8 & 22.0 & 52.8 & 27.5 & 34.9 & 54.4 & 11.3 & 34.0 & 32.7 & 7.1 & 16.4 & 35.3 & 76.1 & 47.5 & 39.6\\
& CMF & Ensemble & 25.5 & 59.3 & 41.0 & 36.3 & 11.4 & 47.1 & 20.3 & \textbf{32.8} & 48.8 & \textbf{7.7} & 28.3 & 28.1 & \textbf{5.5} & 8.1 & 25.7 & 74.0 & \textbf{40.7} & 31.8\\
& ROID & Ensemble & \textbf{24.1} & \textbf{58.4} & \textbf{39.9} & \textbf{36.2} & \textbf{10.4} & \textbf{45.9} & \textbf{18.8} & 33.2 & \textbf{48.6} & 7.9 & \textbf{28.1} & \textbf{28.0} & \textbf{5.5} & \textbf{7.2} & \textbf{25.2} & \textbf{73.9} & 41.4 & \textbf{31.3}\\
\hline\hline
\multirow{7}{*}{\rotatebox[origin=c]{90}{ViT-L-14}} & Source & Ensemble & 24.5 & 58.6 & 29.3 & 30.1 & 12.2 & 40.4 & 22.4 & 24.4 & 43.3 & 7.0 & 22.1 & 25.0 & 5.5 & 10.8 & 30.9 & 68.1 & 39.3 & 29.1\\ 
\cline{2-21}
& TENT & Ensemble & 24.5 & 53.5 & 29.2 & 30.2 & 12.1 & 40.1 & 22.1 & 24.4 & 43.4 & 7.0 & 22.1 & 24.9 & 5.6 & 10.8 & 30.9 & 68.1 & 36.1 & 28.5\\
& ETA & Ensemble & 24.6 & 77.7 & 29.0 & 30.3 & 12.1 & 39.8 & 40.5 & 24.4 & 43.4 & 6.9 & 21.9 & 24.9 & 5.5 & 10.9 & 30.6 & 68.0 & 39.3 & 31.2\\
& SAR & Ensemble & 27.2 & 60.7 & 29.1 & 30.3 & 12.2 & 44.1 & 22.4 & 24.3 & 43.6 & 6.9 & 23.0 & 25.0 & 5.7 & 11.1 & 31.6 & 67.9 & 42.7 & 29.9\\
& DeYO & Ensemble & 24.6 & 55.9 & 28.8 & 30.6 & 11.8 & 39.8 & 21.4 & 24.2 & 43.3 & 6.9 & 22.0 & 25.0 & 5.7 & 11.1 & 31.0 & 67.9 & 40.3 & 28.8\\
& CMF & Ensemble & 18.7 & 41.8 & \textbf{19.6} & 28.3 & 5.2 & 34.3 & 15.3 & \textbf{22.8} & \textbf{36.6} & \textbf{3.8} & \textbf{18.4} & \textbf{21.7} & \textbf{4.5} & 5.5 & 23.3 & 64.7 & \textbf{34.6} & \textbf{23.5}\\
& ROID & Ensemble & \textbf{17.6} & \textbf{41.7} & 19.7 & \textbf{28.1} & \textbf{4.9} & \textbf{33.3} & \textbf{15.1} & 22.9 & 36.8 & \textbf{3.8} & \textbf{18.4} & 21.8 & \textbf{4.5} & \textbf{5.1} & \textbf{22.8} & \textbf{64.4} & 39.1 & \textbf{23.5}\\
    \hline
\end{tabular}}
\label{tab:tta_correlated}
\end{table}